\documentclass{article} 
\usepackage{iclr2025_conference,times}

\usepackage{amsmath,amsfonts,bm}

\def\eqref#1{equation~\ref{#1}}

\def\1{\bm{1}}

\DeclareMathAlphabet{\mathsfit}{\encodingdefault}{\sfdefault}{m}{sl}
\SetMathAlphabet{\mathsfit}{bold}{\encodingdefault}{\sfdefault}{bx}{n}

\usepackage{hyperref}
\usepackage{url}
\usepackage{graphicx}
\usepackage{wrapfig}
\usepackage{multirow}
\usepackage{booktabs}
\usepackage{cleveref}
\usepackage{tabularx}

\title{Generalizable Non-Line-of-Sight Imaging with Learnable Physical Priors}

\author{Shida Sun, Yue Li, Yueyi Zhang \& Zhiwei Xiong \\
University of Science and Technology of China\\
}

\iclrfinalcopy 
\begin{document}

\maketitle

\begin{abstract}
Non-line-of-sight (NLOS) imaging, recovering the hidden volume from indirect reflections, has attracted increasing attention due to its potential applications. Despite promising results, existing NLOS reconstruction approaches are constrained by the reliance on empirical physical priors, e.g., single fixed path compensation. Moreover, these approaches still possess limited generalization ability, particularly when dealing with scenes at a low signal-to-noise ratio (SNR). To overcome the above problems, we introduce a novel learning-based solution, comprising two key designs: Learnable Path Compensation (LPC) and Adaptive Phasor Field (APF). The LPC applies tailored path compensation coefficients to adapt to different objects in the scene, effectively reducing light wave attenuation, especially in distant regions. Meanwhile, the APF learns the precise Gaussian window of the illumination function for the phasor field, dynamically selecting the relevant spectrum band of the transient measurement. Experimental validations demonstrate that our proposed approach, only trained on synthetic data, exhibits the capability to seamlessly generalize across various real-world datasets captured by different imaging systems and characterized by low SNRs.
\end{abstract}

\section{Introduction}
\label{sec:intro}
\begin{figure}[tb]
  \centering
  \includegraphics[width=120mm]{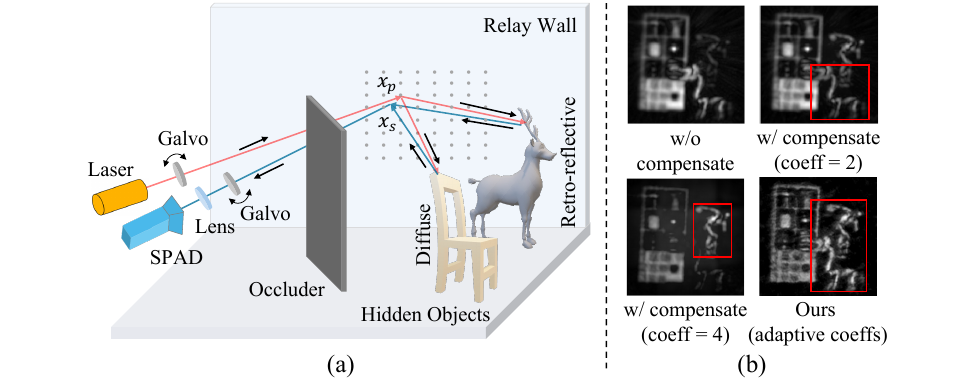}
  \vspace{-0.1cm}
  \caption{(a) An overview of the NLOS imaging system, including objects with distinct surface materials. (b) Reconstructed images from our method and RSD~\citep{liu2019non} with different compensation coefficients. Near to Far: Dragon, Bookshelf, Statue.
  }
  \vspace{-0.8cm}
  \label{fig:nlos_setting}
\end{figure}

Non-line-of-sight (NLOS) imaging represents a groundbreaking advancement in visual perception, enabling the visualization of hidden objects with significant implications in diverse fields, including autonomous navigation, remote sensing, disaster recovery, and medical diagnostics~\citep{bauer2015non, lindell2019acoustic, scheiner2020seeing, laurenzis2017dual, wu2021non, maeda2019recent}.
By harnessing sophisticated time-of-flight (ToF) configuration, NLOS imaging systems can effectively capture light signals bounced off hidden objects, even when direct line-of-sight visibility is obstructed, as illustrated in Fig. \ref{fig:nlos_setting}(a). The core components of such systems typically include pulse lasers, which emit short bursts of light, and time-resolved detection sensors like Single Photon Avalanche Diode (SPAD) and Time-Correlated Single Photon Counting, which precisely capture the flight of the time that takes for photons to travel from the light source to the hidden object and back to the SPAD. The captured signals, known as transient measurement, undergo reconstruction using various algorithms, including traditional approaches~\citep{velten2012recovering, arellano2017fast, liu2019non} and learning-based approaches~\citep{chen2020learned, grau2020deep, mu2022physics, yu2023enhancing, li2023nlost, li2024deep}.

For the traditional approaches, the back projection algorithms~\citep{laurenzis2013non, velten2012recovering} and the light path transport algorithms \citep{heide2019non, o2018confocal}
typically assume isotropically scattering, no inter-reflection, and no occlusions within the hidden scenes. However, these approaches always yield noisy results and lack details. Conversely, the wave propagation approaches~\citep{lindell2019wave,liu2020phasor} require no special assumptions and tend to produce better results while being sensitive to scenes with large depth variations. 
Learning-based approaches~\citep{chen2020learned,mu2022physics,li2023nlost} leverage the powerful representation capabilities of neural networks, and push NLOS reconstruction to a higher level.

Despite promising results, current NLOS reconstruction algorithms are constrained by the reliance on empirical physical priors and are still confronted with challenges. 
The primary challenge is Radiometric Intensity Fall-off (RIF), i.e., the intensity of the reflected photons attenuates and the degree of attenuation is related to the surface material of the hidden object.

To address this phenomenon, quadratic and quartic operations are commonly applied to the light propagation path for retro-reflective and diffuse surfaces, respectively \citep{o2018confocal}, to compensate for intensity attenuation. 
However, since various surface materials coexist within the same scene, applying path compensation based on a single material type across the entire scene, as performed in physical priors or previous work, may not effectively counteract the effects of attenuation.
Additionally, the problem is exacerbated by the low quantum efficiency of the imaging system, particularly over long distances. 
As shown in Fig. \ref{fig:nlos_setting}(b), using a single coefficient to compensate for the entire scene can enhance the reconstruction of objects with corresponding material properties, but it will significantly reduce the SNR for other objects in the same scene.
Another challenge is the limited generalization ability mainly caused by various noises. In this study, we concentrate on two specific noise sources: the dark count of the SPAD and the ambient light~\citep{hernandez2017computational}. As the data acquisition time decreases, the signal-to-noise ratio (SNR) decreases, resulting in higher noise levels.
The Poisson-distributed noise photons degrade the quality of transient measurements especially at low SNR, manifesting as high-frequency aliasing. This phenomenon poses grave challenges to existing approaches, with traditional ones yielding a plethora of artifacts, and learning-based ones experiencing a breakdown in their ability to generalize. 

To address the above two challenges, we propose a novel learning-based approach by leveraging the virtual wave phasor field~\citep{liu2019non}. Our approach incorporates two key designs: the Learnable Path Compensation (LPC) and the Adaptive Phasor Field (APF).
Given that reflected light with different degrees of RIF may be captured simultaneously, the LPC utilizes three physics-based predefined compensation weights to initialize the features of transient measurements for path compensation.
Subsequently, a convolutional neural network is trained to implicitly learn and assign distinct compensation coefficients to each scanning point in the transient measurements. By utilizing these learnable compensation coefficients, the LPC adaptively mitigates light wave attenuation in the same scene, as shown in Fig. \ref{fig:nlos_setting}(b), particularly for distant regions.
Meanwhile, the APF learns an applicable standard deviation for the Gaussian window of the illumination function, allowing it to dynamically choose the relevant spectrum band for each transient measurement. The emphasis on the effective spectrum enables the discrimination of useful information from noise under distinct SNR conditions.

To demonstrate the efficacy of our proposed approach, we train the approach on a synthetic dataset and subsequently test them on unseen data, including both synthetic and real-world datasets captured from different imaging systems. The exceptional performance on unseen synthetic data and the diverse real-world data highlight the robust generalization capabilities of our approach. Even under challenging conditions, i.e., fast acquisition time and low SNR, our method consistently outperforms competitors. To further increase the diversity of NLOS data, we provide three real-world data captured by our own NLOS imaging system to conduct more comprehensive experiments.

In summary, the contributions of this paper can be listed as follows:
\begin{itemize}
    \item We propose a novel learning-based solution for NLOS reconstruction, breaking the reliance on empirical physical priors and boosting the generalization capability.
    \item We design the LPC to adaptively mitigate the light attenuation in the same scene. The embedded learnable physical prior greatly improves the generalization capability across different object materials, especially for long-distance regions.
    \item We design the APF to prioritize the relevant information from the frequency domain, which improves the generalization capability across transient measurements under distinct SNR conditions.
    \item Our proposed approach, trained on synthetic data, achieves the best generalization performance on both synthetic and publicly real-world datasets with diverse SNRs. Additional real-world data captured by our own imaging system further showcases the capability of our approach.
\end{itemize}

\section{Related Work}

\subsection{Traditional Approaches}
In the rapidly advancing field of NLOS imaging, significant progress has been made towards unveiling hidden objects. The groundwork was established by \cite{kirmani2009looking}, who pioneered the use of time-resolved imaging to navigate photons around obstructions, despite facing computational challenges due to complex multi-path light transport. Efforts to streamline the complex inverse problem have led to the development of back projection approaches, notable for their ability to approximate the geometry of obscured objects through ultrafast time-of-flight information capturing and light geometric relationship~\citep{velten2012recovering,arellano2017fast}. The Light-cone Transform (LCT), marked by the introduction of simple assumptions for light propagation, further facilitated the NLOS reconstruction with unprecedented detail by solving inverse problems in the linear space~\citep{o2018confocal}. The wave propagation approaches like frequency-wavenumber migration (FK)~\citep{lindell2019wave} and Rayleigh Sommerfeld Diffraction (RSD)~\citep{liu2020phasor,liu2019non} provided enhanced accuracy for NLOS imaging by considering the interaction between the light wave and multiple hidden object surfaces. Despite considerable progress, traditional algorithms are still limited with challenges in noise effects and complicated scenes.
\vspace{-0.2cm}

\subsection{Learning-based Approaches}
Recently, learning-based approaches have been gradually introduced into NLOS imaging. \cite{grau2020deep} proposed the first end-to-end learnable network for NLOS reconstruction. The UNet~\citep{ronneberger2015u} based network regressed the depth from transient measurements directly. However, it is an unstable solution that transforms the non-linear spatial-temporal domain into the linear spatial domain solely by convolution layers. The instability is particularly evident in real-world scenarios, resulting in poor reconstructions. To solve this problem, \cite{chen2020learned} developed the physics-based feature propagation module (LFE, Learned Feature Embeddings) to transform different domains, narrowing the domain gap between the synthetic and real-world data. Building on the insights from NeRF~\citep{mildenhall2021nerf}, recent solutions~\citep{mildenhall2021nerf,mu2022physics} can render the albedo of hidden objects through the radiance field in the unsupervised manner, which consumes large computation time for each inference. Through analysis of transients histogram, \cite{li2023nlost} produced the first transformer-based framework (NLOST) for capturing local and global correlations, while entailing a substantial computational burden. \cite{yu2023enhancing} introduced a learnable Inverse Kernel (I-K) with attention mechanisms. However, I-K is actually tailored for the point spread function of the imaging system rather than the transient measurements. While the above physics-based approaches~\citep{chen2020learned,li2023nlost,yu2023enhancing} consistently improve NLOS reconstruction performance, they still encounter challenges when reconstructing real-world scenes with diverse object materials. Additionally, these approaches overlook the generalization of the real-world transient measurements with low SNRs. In this paper, we present specific solutions tailored to these two challenges.

\section{Methodology}
\subsection{Imaging Formulation}
\label{sec:nlos_imaging_model}
\begin{figure}[tb]
  \centering
  \includegraphics[width=1.0\textwidth]{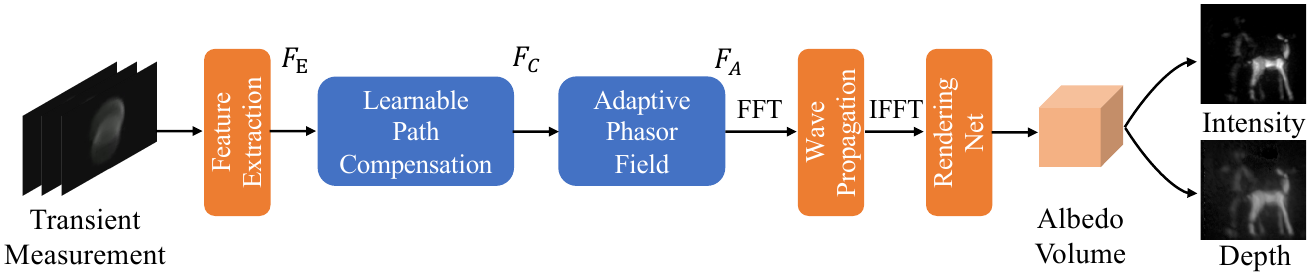}
  \vspace{-0.55cm}
  \caption{An overview of our proposed approach. Given the transient measurements as input, the approach generates the albedo volume, intensity image, and depth map.
  }
  \vspace{-0.2cm}
  \label{fig:whole_pipeline}
\end{figure}
We begin with an impulse response captured from the relay wall, noted as $H(x_p \to x_s,t)$. With the virtual illumination source wavefront $\mathcal{P}(x_p,t)$, the phasor field at the virtual aperture $\mathcal{P}(x_s, t)$ can be formulated~\citep{liu2019non, liu2020phasor} as:
\abovedisplayskip=0pt 
\abovedisplayshortskip=0pt 
\belowdisplayskip=0pt  
\belowdisplayshortskip=0pt 
\begin{align}
      \mathcal{P}(x_s, t) =  \int_{P} \mathcal{P}(x_p,t)*\left(\frac{1}{r^{z}}\cdot H(x_p \to x_s,t)\right) dx_p ,
    \label{equ:nlos_1_model}
\end{align}
where $*$ denotes the convolution operator, and $x_p$ and $x_s$ represent the illumination point and the scanning point, respectively. The term $1/r^{z}$ represents the RIF, where $r$ is the distance between the scanning point and the target point. The parameter $z$, which indicates the attenuation coefficient associated with different surface materials, is the parameter the LPC module is designed to learn.

The $\mathcal{P}(x_p,t)$, referred to as the illumination function, is defined as a Gaussian-shaped function modulated with the virtual wave $e^{j\Omega_Ct}$, which can be represented as illumination phasor field $\mathcal{P_{\mathcal{F}}}(x_p, \Omega)$ in the Fourier domain as~\cite{liu2019non}:
\begin{align}
    \mathcal{P_{\mathcal{F}}}(x_p, \Omega) = \delta(x_p-x_{vp}) \cdot 
    \bigg( 2\pi\delta(\Omega-\Omega_C) \underset{\mathcal{F}}{*} \sigma\sqrt{2\pi} 
    \exp\left(-\frac{\sigma^2\Omega^2}{2}\right) \bigg),
    \label{equ:ipf}
\end{align}
where $\mathcal{F}$ represents the Fourier domain, $x_{vp}$ denotes the position at the virtual light source, $\delta$ is the Dirac function, $\Omega_C$ denotes the central frequency of the wave, and $\sigma$ represents the standard deviation. The standard deviation of a Gaussian is inversely proportional to its pass-band width in the frequency domain, which can be learned and adjusted automatically by our APF module.

The point $I(x,y)$ of the hidden object can be reconstructed from $\mathcal{P}(x_s, t)$ with the wave propagation function $\Phi(\cdot)$, which is modeled by the Rayleigh-Sommerfeld Diffraction integral:
\begin{align}
    I(x,y)= \Phi \left ( \mathcal{P}(x_s, t) \right ) .
    \label{equ:nlos_2_model}
\end{align}
Without loss of generality, considering Poisson noise resulting from ambient light and background noise, the computational model of SPAD sensor~\citep{saunders2019computational,grau2020deep} can be written as:
\begin{align}
    H'(x_p \to x_s,t) \sim \text {Poisson}(H(x_p \to x_s,t) + B) ,
    \label{equ:noise_model}
 \end{align}
where $B$ represents detected photons from background noise and dark counts~\citep{bronzi2015spad} of SPAD sensors. Poisson($\cdot$) represents the Poisson distribution~\citep{snyder2012random}.
\subsection{Overview}
\begin{figure}[tb]
  \centering
  \includegraphics[width=\textwidth]{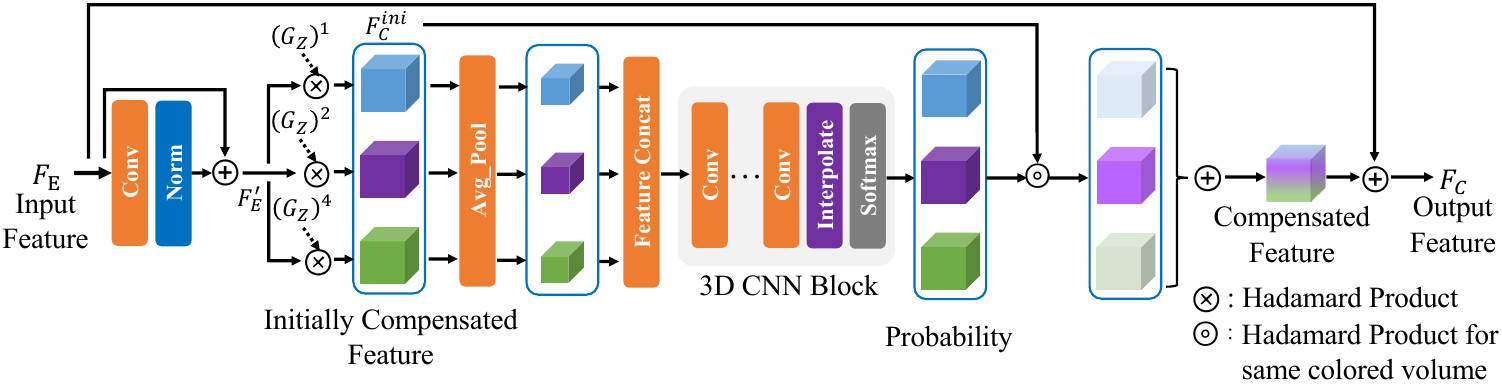}
  \vspace{-0.7cm}
  \caption{The pipeline of the LPC.
  }
  \vspace{-0.6cm}
  \label{fig:LDC_pipeline}
\end{figure}
To address the problems mentioned in Section \ref{sec:intro}, we integrate the proposed LPC and APF modules into the LFE~\citep{chen2020learned} framework, which comprises a feature extraction module, a wave propagation module, and a rendering module. An overview of the network is shown in Fig. \ref{fig:whole_pipeline}. Given transient measurements as input, similar to those described in the literature~\citep{chen2020learned, li2023nlost}), the feature extraction module downsamples the transient measurements in both spatial and temporal dimensions and extracts feature embeddings $F_E$.

Instead of directly applying the wave propagation module to convert transient measurements to the spatial domain, we first employ the LPC to learn different attenuation coefficients for each scanning position at the aperture. This allows us to compute the corresponding feature compensation amplitudes, resulting in the compensated feature $F_C$. Subsequently, the APF module predicts the optimal frequency domain window width for the illumination function, which illuminates $F_C$ and generates $F_A$. Finally, the wave propagation and rendering module converts $F_A$ from the spatial-temporal domain to the spatial domain and renders intensity and depth images. We provide details of the network in the Supplementary Material.
\subsection{Learning to Compensate Radiometric Intensity \hspace{0pt}Fall-off}
To alleviate the aforementioned RIF, we design the LPC module, which can predict the clean transient measurements before attenuation. An overview of the LPC is shown in Fig. \ref{fig:LDC_pipeline}. Given the features $F_E$ from the previous feature extraction module, the LPC first enhances the features using a convolutional layer with normalization, yielding $F_{E}^{'}$. Let $G_Z$ denote the grid representing the distance from the hidden volume to the relay wall, we predefine three path compensation weights $\{(G_Z)^r, r=1, 2, 4\}$, which correspond to different attenuation amplitudes of surface materials, as referenced in~\cite{o2018confocal} and~\cite{liu2020phasor}. The weights and enhanced features are multiplied to obtain initially compensated features $F_{C}^{ini}$, which can be expressed as:
\begin{align}
    F_{C}^{ini} = \left \{(G_{Z})^1, (G_{Z})^2, (G_{Z})^4 \right \} \otimes F_{E}^{'} ,
\end{align}
where $\otimes$ denotes the Hadamard product.

After that, the initial compensated features are down-sampled across the spatial dimensions using an average pooling layer. Instead of predicting the RIF term directly, we design the LPC to predict probabilities of initial compensation features first, and then the weights and features are combined through a weighted sum. In such a way, the LPC is capable of explicitly selecting appropriate compensation amplitudes based on physical constraints. The downsampled features thus undergo a series of operations including convolution layers, interpolation, and the Softmax operation, which outputs probabilities. The probabilities and the initial compensated features are then multiplied using the Hadamard product, resulting in compensated features. Subsequently, the compensated features and the input features are added together, outputting the final compensated features. 

As demonstrated in Section \ref{sec:ablation_modules}, our carefully designed LPC module effectively mitigates the RIF issue, enhancing the reconstruction performance for challenging real-world scenes, especially in complex and distant regions.

\subsection{Denoising with Adaptive Phasor Field}
\begin{figure}[tb]
  \centering
  \includegraphics[width=\textwidth]{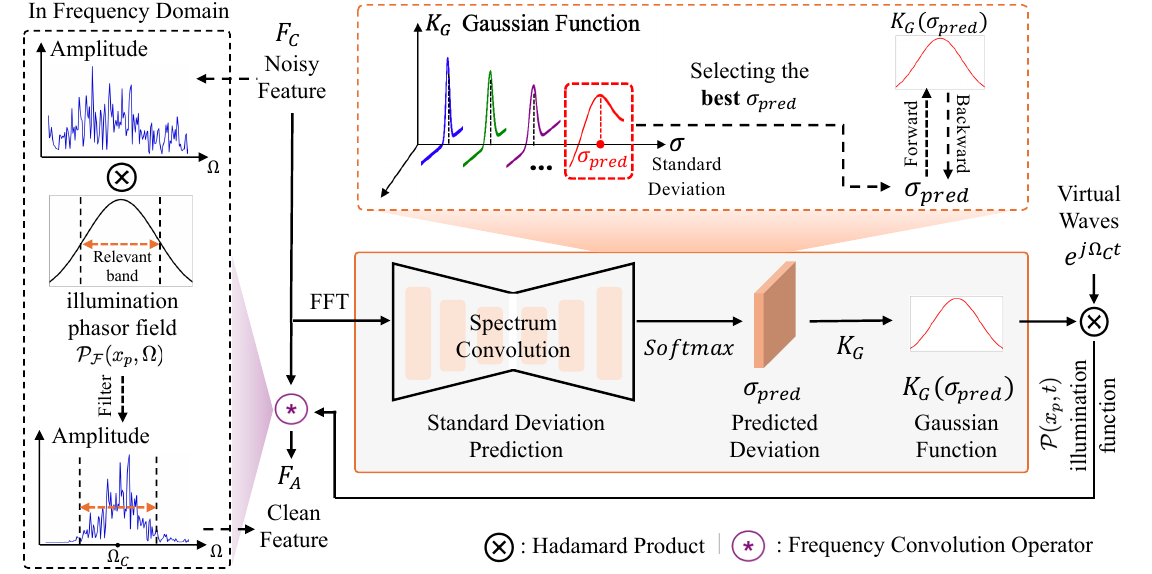}
  \vspace{-0.5cm}
  \caption{The pipeline of the APF. The module predicts the illumination function with an appropriate bandwidth to compensate for the noisy transient features, outputting clean, denoised features.
  }
  \vspace{-0.5cm}
  \label{fig:APF_pipeline}
\end{figure}
As described in the imaging formulation in Section \ref{sec:nlos_imaging_model}, the transient measurement is illuminated by the virtual illumination function. In the frequency domain, the illumination phasor field $\mathcal{P_{\mathcal{F}}}(x_p, \Omega)$ acts as a Gaussian filter on the feature of transient measurements, modulated to the central frequency $\Omega_C$. It should be noted that not all frequency components contribute positively to the final scene reconstruction, including frequency components associated with noise~\citep{liu2020phasor, hernandez2017computational}. Applying an illumination function to the feature of transient measurements can be understood as a process of selecting a certain effective frequency spectrum band $\bigtriangleup\Omega$. The bandwidth of the Gaussian function for the illumination function is decided by the standard deviation, which can be expressed as
\begin{equation}
    \bigtriangleup\Omega = \frac{1}{2\pi\sigma}.
\end{equation}
For convenience, the $\bigtriangleup\Omega$ is defined as the 3 dB bandwidth. Selecting an appropriate standard deviation is crucial for obtaining clean measurements. However, past works have relied on a single empirical standard deviation, which is not conducive to selecting the correct frequency components for the reconstruction of complicated scenarios.

To address this problem, we devise the APF module to adaptively learn the standard deviation, as illustrated in Fig. \ref{fig:APF_pipeline}.
Given the feature $F_{C}$, the first step is to transform the feature into the frequency domain along the temporal dimension. This allows the module to intuitively learn to distinguish between useful information and noise directly in the frequency domain. Subsequently, the Fourier features are convolved across the spatial and the spectrum parts successively to further enhance features. We then employ additional fully connected layers to predict the standard deviations $\sigma_{pred}$ from frequency feature representation, generating the adaptive Gaussian function $K_{G}(\sigma)$ in the frequency domain. As such, the illumination phasor field can be formulated by the adaptive Gaussian function and the virtual waves $e^{j\Omega_Ct}$, as
\begin{equation}
\vspace{-0.2cm}
    \mathcal{P_{\mathcal{F}}}(x_p, \Omega) = \delta(x_p-x_{vp}) \cdot 
    \bigg( \mathcal{F}\left(e^{j\Omega_Ct}\right) \underset{\mathcal{F}}{*} K_{G}(\sigma_{pred}) \bigg),
\end{equation}
where
\begin{equation}
     K_{G}(\sigma) = \sigma\sqrt{2\pi} \exp\left(-\frac{\sigma^2\Omega^2}{2}\right).
\end{equation}
Finally, the input features $F_{C}$ and the illumination phasor field are convolved across the temporal dimension, as
\begin{equation}
     F_A = F_C * \mathcal{P_{\mathcal{F}}}(x_p, \Omega) = \mathcal{F}^{-1}\bigg(\mathcal{F}(F_C) \cdot \mathcal{F}\left(\mathcal{P}(x_p, t)\right)\bigg),
\end{equation}
where $F_A$ is the output feature in the temporal domain at the scanning point, and $*$ means the convolution operator.

As demonstrated in Section \ref{sec:real-world results} and Section \ref{sec:ablation_modules}, the APF module selectively emphasizes useful information and attenuates noise across various SNR conditions within the transient measurements, thereby boosting the generalization capability and improving the reconstruction quality.

\subsection{Loss Function}
\label{sec:network_training}
The approach is trained in an end-to-end manner. The total loss consists of the intensity loss and the depth loss, balanced by a regularization weight $\lambda$: 
\begin{align}
    \mathcal{L} = \mathcal{L_I}(I, \hat{I}) + \lambda\mathcal{L_D}(D, \hat{D}) ,
\end{align}
and
\begin{align}
    \mathcal{L_I}(I, \hat{I}) = \frac{1}{N}\sum_{i}^{N}(I_i-\hat{I}_i)^2, 
    \mathcal{L_D}(D, \hat{D}) = \frac{1}{N}\sum_{i}^{N}(D_i-\hat{D}_i)^2,
\end{align}
where $\hat{I}$ and $I$ denote the reconstructed intensity image and the ground truth, respectively. $\hat{D}$ and $D$ denote the recovered depth map and corresponding ground truth. $N$ denotes the total number of pixels of the intensity image and depth map.
\section{Experimental Results}
\label{sec:results}
\begin{table}[!t]
  \centering
  \vspace{-0.25cm}
  \caption{Quantitative comparisons of different approaches upon the \textbf{Seen} test set. The best in \textbf{bold}, the second in underline.}
  \vspace{0.3cm}
  \label{tab:syn_result}
  \footnotesize
  \begin{tabularx}{\linewidth}{cccXcXcX}
    \toprule
    \multirow{2.5}{*}{\textbf{Method}} & \multirow{2.5}{*}{\textbf{Backbone}} & \multirow{2.5}{*}{\textbf{Memory}} & \multirow{2.5}{*}{\textbf{Time}} & \multicolumn{2}{c}{\phantom{a}\textbf{Intensity}} & \multicolumn{2}{c}{\textbf{Depth}} \\
    \cmidrule(lr){5-6} \cmidrule(lr){7-8}
    &  &  &  & PSNR$\uparrow$ & SSIM$\uparrow$ & RMSE$\downarrow$ & MAD$\downarrow$ \\
    \midrule
    LCT \citep{o2018confocal} & Physics & 18 GB & 0.11 s & 19.51 & 0.3615 & 0.4886 & 0.4639 \\
    FK \citep{lindell2019wave} & Physics & 26 GB & 0.16 s & 21.69 & 0.6283 & 0.6072 & 0.5801 \\
    RSD \citep{liu2019non} & Physics & 33 GB & 0.23 s & 21.74 & 0.1817 & 0.5677 & 0.5320 \\
    LFE \citep{chen2020learned} & CNN & 13 GB & 0.05 s & 23.27 & 0.8118 & 0.1037 & 0.0488 \\
    I-K \citep{yu2023enhancing} & CNN & 14 GB & 0.08 s & 23.44 & \underline{0.8514} & 0.1041 & 0.0476 \\
    NLOST \citep{li2023nlost} & Transformer & 38 GB & 0.38 s & \underline{23.74} & 0.8398 & \underline{0.0902} & \underline{0.0342} \\
    Ours & CNN & 17 GB & 0.24 s & \textbf{23.99} & \textbf{0.8703} & \textbf{0.0874} & \textbf{0.0312} \\
    \bottomrule
  \end{tabularx}
  \vspace{-0.29cm}
\end{table}
\subsection{Baselines and Datasets}
\textbf{Baselines selection}. To assess the efficacy of our proposed approach, we undertake thorough validations by comparing it against several baseline approaches on the synthetic and real-world datasets. These baselines encompass three traditional approaches commonly used in the field: LCT \citep{o2018confocal}, FK \citep{lindell2019wave}, and RSD \citep{liu2019non}, as well as three learning-based approaches: LFE \citep{chen2020learned}, I-K \citep{yu2023enhancing}, and NLOST \citep{li2023nlost}.
 
\textbf{Public data}. For the synthetic dataset, we utilize a publicly available dataset generated from LFE~\citep{chen2020learned}. A total of 2704 samples are used for training and 297 samples for testing, denoted as \textbf{Seen} test set. Each transient measurement possesses a resolution of 256$\times$256$\times$512, with a bin width of 33 ps and a scanning area of 2m$\times$2m. To assess the generalization capabilities, we rendered 500 transient measurements from the objects not included in the Seen test set, denoted as \textbf{Unseen} test set. For qualitative validation, particularly in complicated scenarios, we employ publicly available real-world data from FK~\citep{lindell2019wave} and also the data from NLOST~\citep{li2023nlost} with low SNR conditions. For example, instead of the commonly used measurements with 180 minutes acquisition time, we utilize the measurements with 10 minutes acquisition time in FK~\citep{lindell2019wave}. We preprocess the real-world data for testing, and the real-world data has a spatial resolution of 256$\times$256 and a bin width of 32ps. 
 
\textbf{Self-captured data}. To further increase the diversity of NLOS data, we also capture additional real-world measurements using our own active confocal imaging system. The system utilizes a 532 nm VisUV-532 laser that generates pulses with an 85 picosecond width and a 20 MHz repetition rate, delivering an average power output of 750 mW. The laser pulses are directed onto the relay wall using a two-axis raster-scanning Galvo mirror (Thorlabs GVS212). Both the directly reflected and diffusely scattered photons are then collected by another two-axis Galvo mirror, which funnels them into a multimode optical fiber. This fiber channels the photons into a SPAD detector (PD-100-CTE-FC) with approximately 45\% detection efficiency. The motion of both Galvo mirrors is synchronized and controlled via a National Instruments acquisition device (NI-DAQ USB-6343). The TCSPC (Time Tagger Ultra) records the pixel trigger signals from the DAQ, synchronization signals from the laser, and photon detection signals from the SPAD. The overall system achieves a temporal resolution of around 95 ps. During data acquisition, the illuminated and sampling points remain aligned in the same direction but are intentionally offset to prevent interference from directly reflected photons. As such setting, we capture three transient measurements from customized scenes, each containing different types of surface materials. All measurements are captured over a duration of 10 minutes.
\subsection{Implementation Details and Metrics}
We implement our approach using the PyTorch framework \citep{paszke2019pytorch}. For optimization, we employ the Adam optimizer \citep{kingma2014adam} with a learning rate of $6\times10^{-5}$ and a weight decay of 0.95. The $\lambda$ is set to 1. Baseline approaches are implemented using their respective public code repositories. The batch size is uniformly set to 1 for all approaches. Training is conducted for 50 epochs using a single NVIDIA RTX 3090 GPU, except for NLOST, which is trained on Tesla A100 GPUs. Due to memory consumption, NLOST is trained on transient measurements with the shape of $128\times128\times512$, and the results are interpolated to $256\times256$ for comparison.

For quantitative evaluation in intensity reconstruction, we adopt peak signal-to-noise ratio (PSNR) and structural similarity metrics (SSIM) averaged on the test set. For depth reconstruction, we compute the root mean square error (RMSE) and mean absolute distance (MAD) for test samples. Following \cite{li2023nlost}, we crop the central region for a more reliable evaluation.

\subsection{Comparison on Synthetic Data}
\label{sec:comparison_on_syn_data}
\begin{figure*}[!t]
  \centering
  \includegraphics[width=1.0\textwidth]{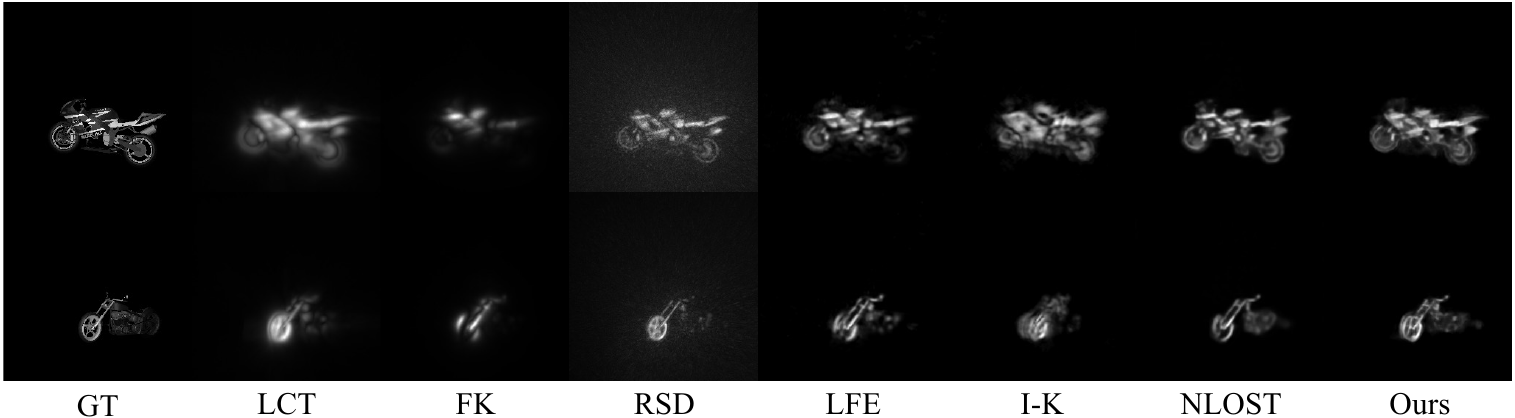}
  \vspace{-0.5cm}
  \caption{Intensity results recovered by different approaches on the \textbf{Seen} test set. GT means ground truth of the intensity images.
  }
  \vspace{-0.3cm}
  \label{fig:syn_result}
\end{figure*}

\textbf{Quantitative evaluation}. The quantitative evaluations presented in Table \ref{tab:syn_result} demonstrate that our approach achieves decent advancements in NLOS reconstruction. For the synthetic results, our approach outperforms all competitors in terms of all evaluation metrics. Specifically, our approach exhibits a substantial enhancement over traditional approaches, achieving a 2.25 dB increase in PSNR compared to the leading approach RSD. Furthermore, when compared with the recent state-of-the-art (SOTA) learning-based approaches I-K and NLOST, our approach still achieves a 0.55 dB and 0.25 dB improvement in PSNR, respectively. The merits of our approach are further substantiated by the highest SSIM for intensity, which underscores the superior capability of our network in preserving the structural integrity of hidden scenes. Additionally, for the depth estimation, our approach reduces the RMSE and MAD metrics by 3.10\% and 8.77\%, respectively, over the strongest competitor NLOST.

Notably, the existing Transformer-based SOTA approach NLOST requires approximately 38 GB of GPU memory and a substantial amount of inference time. In contrast, our approach achieves higher performance while using only half the memory and requiring less inference time.

\begin{figure*}[!t]
  \centering
  \includegraphics[width=1.0\textwidth]{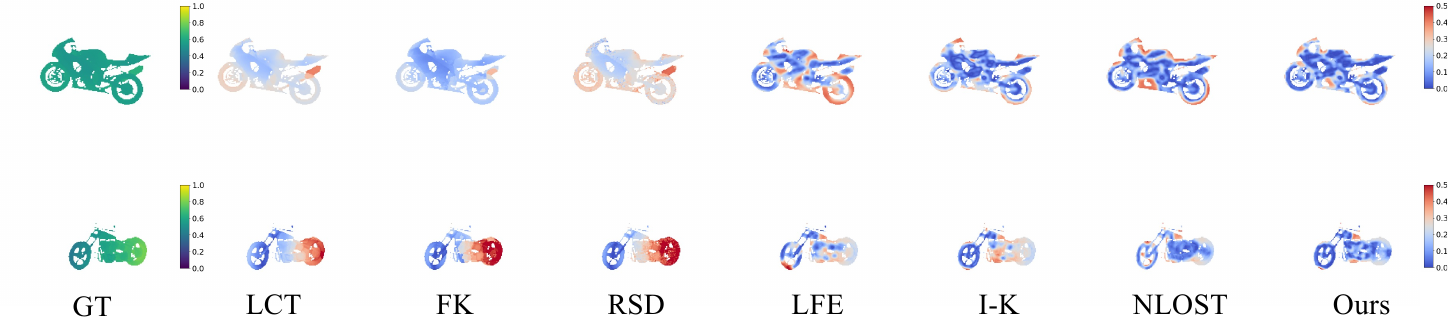}
  \vspace{-0.4cm}
  \caption{Depth error maps from different approaches on the \textbf{Seen} test set. The first column denotes the ground-truth depth map, and the other columns indicate the depth error maps. The color bars show the value of depth and error maps, respectively.
  }
  \vspace{-0.5cm}
  \label{fig:syn_dep_result}
\end{figure*}

\textbf{Qualitative evaluation}. We present the qualitative results of intensity images and depth error maps for visualization comparisons, depicted in Fig. \ref{fig:syn_result} and Fig. \ref{fig:syn_dep_result}. Regarding the intensity visualization comparisons,  LCT reconstructs the main content yet sacrifices details, FK fails to recover most of the structural information, and RSD introduces significant noise in the background. The LFE and I-K perform better than traditional approaches but still lack details. Compared to the SOTA approach NLOST, our approach generates content with greater fidelity and high-frequency details (e.g., the texture of the scene in the first row). In terms of the depth error map, the blue regions dominate the scene in the error map corresponding to our approach, indicating the smallest magnitude of the error. In contrast, traditional approaches as well as LFE demonstrate a greater tendency for errors, as shown by the increased presence of red parts, especially in distant regions (e.g., the right part of the motorcycles in the second row). These areas are challenging due to the complex geometrical features and distinct RIF degrees with different kinds of materials. While I-K and NLOST show improvement over the former approaches, they still fail to precisely estimate the depth in the wheel area, where our approach succeeds.

\begin{figure*}[t]
  \centering
  \includegraphics[width=1.0\textwidth]{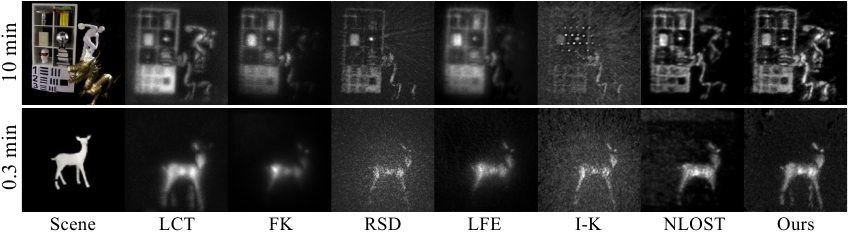}
  \vspace{-0.5cm}
  \caption{Visualization comparison on the public real-world data~\citep{lindell2019wave, li2023nlost}. The left annotation indicates the shortest acquisition time in total. Zoom in for details.
  }
  \vspace{-0.1cm}
  \label{fig:real_world_1}
\end{figure*}
\begin{figure}[t]
  \centering
  \includegraphics[width=1.0\textwidth]{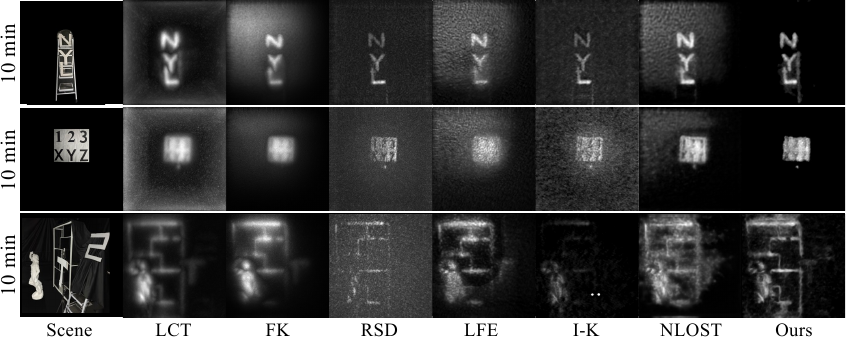}
  \vspace{-0.5cm}
  \caption{Visualization comparison on our self-captured real-world data. The left annotation indicates the total acquisition time. Zoom in for details.
  }
  \vspace{-0.5cm}
  \label{fig:real_world_2}
\end{figure}

\textbf{Generalization evaluation}. To further validate the network's generalization performance, we conduct quantitative tests under varying SNR conditions. Specifically, we test different approaches with the \textbf{Unseen} test set under varying SNR levels (10 dB, 5 dB, and 3 dB) of the Poisson noise. Extreme SNR conditions make separating background noise from the limited number of collected photons more challenging, while the new scenes in the Unseen test set validate the performance when transferred to unknown domains. As can be seen in Table \ref{tab:ablation_generalization_snr}, in most cases, our approach achieves the best results compared to other approaches. These outstanding results demonstrate the superior generalization performance of our approach in dealing with test data that is distinct from the training data. This superiority is then further verified on various real-world data that has no ground truth below.

\subsection{Comparison on Real-world Data}
\label{sec:real-world results}

\textbf{Public data}. Results on two public NLOS datasets are presented in Fig. \ref{fig:real_world_1}. When utilizing measurements with reduced acquisition time, nearly all approaches, except for NLOST and our approach, produce reconstructions with significant noise. The traditional approaches, while reconstructing main content, produces blurred results. LFE and I-K manage to reconstruct more objects but struggle to capture high-frequency details. NLOST excels in reducing background noise, but it still misses certain details such as the legs of the deer and the intricate patterns of the tablecloth. Our approach shows remarkable resilience to variation in different acquisition times, consistently delivering detailed reconstructions comparable to those of the same objects captured at high acquisition time. The exceptional robustness of our approach demonstrates the superior generalization ability over the existing approaches.

\textbf{Self-captured data}. Apart from the public data, we also capture several new scenes with our own NLOS system for further assessment. We present results from three distinct scenes: one depicting retro-reflective letters arranged on a ladder (referred to as `ladder'), another featuring a panel composed of multiple A4 sheets inscribed with `123XYZ' (referred to as `resolution'), and the third containing multiple objects with varying surface materials (referred to as `composite'). As shown in Fig. \ref{fig:real_world_2}, it can be observed that learning-based approaches still exhibit less reconstruction noise compared to traditional approaches. In the low SNR scenario of the `ladder', other approaches either fail to reconstruct or produce poor-quality reconstructions. However, our reconstruction exhibits notably high quality, with the ladder legs even discernible. In the heavily attenuated diffuse reflection scenario `resolution', our approach still manages to reconstruct relatively clear details. In the `composite' scene, which includes depth variations and multiple surface materials, our approach produces reconstruction with the least noise and the most complete structural information (e.g., the lower edge of the bookshelf and the letter `S' in the upper right of the scene). The promising outcomes achieved by our approach underscore its superiority over existing approaches.

\begin{table}[t]
  \centering
  \vspace{-0.25cm}
  \caption{Quantitative results on the \textbf{Unseen} test set under different SNRs. The best in \textbf{bold}, the second in underline.}
  \label{tab:ablation_generalization_snr}
  \vspace{0.4cm}
  \scriptsize
  \begin{tabularx}{\linewidth}{c>{\centering\arraybackslash}X>{\centering\arraybackslash}X>{\centering\arraybackslash}X>{\centering\arraybackslash}X>{\centering\arraybackslash}X>{\centering\arraybackslash}X} 
    \toprule
    \multirow{2.5}{*}{\textbf{Method}} & \multicolumn{3}{c}{\textbf{Intensity (PSNR$\uparrow$ / SSIM$\uparrow$)}} & \multicolumn{3}{c}{\textbf{Depth (RMSE$\downarrow$ / MAD$\downarrow$)}} \\
    \cmidrule(lr){2-4} \cmidrule(lr){5-7}
    & 10 dB & 5 dB & 3 dB & 10 dB & 5 dB & 3 dB \\
    \midrule
    LCT & 18.92 / 0.1708 & 18.38 / 0.1195 & 18.06 / 0.1007 & 0.6992 / 0.6499 & 0.7490 / 0.1195 & 0.7666 / 0.7197 \\
    FK & 21.62 / 0.6496 & 21.62 / 0.6471 & 21.62 / 0.6452 & 0.5813 / 0.5562 & 0.5672 / 0.5427 & 0.5598 / 0.5351 \\
    RSD & 22.77 / 0.2045 & 22.48 / 0.1510 & 22.24 / 0.1280 & 0.4198 / 0.3934 & 0.3679 / 0.3358 & 0.3496 / 0.3160 \\
    LFE & 23.22 / 0.8122 & 23.15 / 0.7951 & 23.10 / 0.7805 & 0.1036 / 0.0484 & 0.1041 / 0.0491 & 0.1044 / 0.0496 \\ 
    I-K & 23.45 / \underline{0.8386} & 23.38 / 0.8020 & 23.32 / 0.7689 & 0.1045 / 0.0500 & 0.1071 / 0.0571 & 0.1099 / 0.0636 \\
    NLOST & \underline{23.63} / 0.7747 & \underline{23.74} / \underline{0.8294} & \underline{23.71} / \underline{0.8135} & \underline{0.0939} / \underline{0.0409} & \textbf{0.0909} / \textbf{0.0351} & \underline{0.0918} / \underline{0.0368} \\
    Ours & \textbf{23.91} / \textbf{0.8577} & \textbf{23.83} / \textbf{0.8387} & \textbf{23.80} / \textbf{0.8645} & \textbf{0.0893} / \textbf{0.0333} & \underline{0.0914} / \underline{0.0365} & \textbf{0.0902} / \textbf{0.0332} \\
    \bottomrule
  \end{tabularx}
\vspace{-0.3cm}
\end{table}

\subsection{Ablation Studies}
\label{sec:ablation_modules}
\begin{figure}[th]
  \centering
  \includegraphics[width=1.0\textwidth]{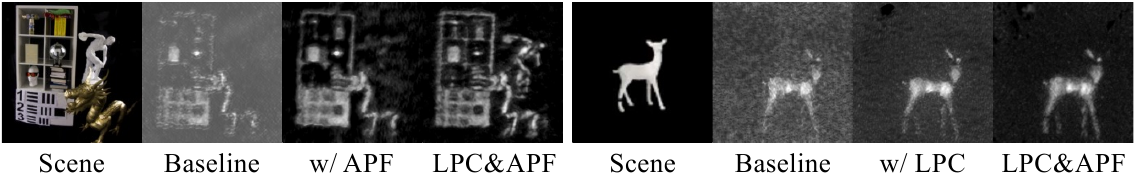}
  \vspace{-0.6cm}
  \caption{Ablation results on public real-world data. Baseline denotes w/o LPC and APF modules. The total acquisition time of the left and right scenes is 10 min and 0.3 min, respectively.
  }
  \vspace{-0.5cm}
  \label{fig:ablation_modules}
\end{figure}

\noindent In this section, we ablate the contribution of the modules. As shown in the qualitative results in Fig. \ref{fig:ablation_modules}, the LPC and the APF modules each contribute to improving the performance of the approach in distinct ways, with their combination yielding the best results. Specifically, it can be seen that the network without the proposed modules loses image details and contains significant noise in the reconstruction. In contrast, introducing the LPC module enhances object details (e.g., the deer's legs), and introducing the APF module suppresses background artifacts. When both the APF and the LPC modules are integrated, the network produces images with complete details and clear boundaries.
\section{Discussion and Conclusion}
In this paper, we propose a novel learning-based approach for NLOS reconstruction including two elaborate designs: learnable path compensation and adaptive phasor field. Experimental results demonstrate that our proposed solution effectively mitigates RIF and improves the generalization capability. Additionally, we contribute three real-world scenes captured by our NLOS imaging system. 
The future work of our study is twofold. We conduct experiments on the confocal imaging system, with the extension to the non-confocal imaging system being one direction of our future research. The modeling of the SPAD acquisition process still exhibits a certain gap from the real-world sensor, and considering additional factors remains a focus for future research.


\bibliography{iclr2025_conference}
\bibliographystyle{iclr2025_conference}


\end{document}